\documentclass[letterpaper, 10 pt, conference]{ieeeconf}  

\IEEEoverridecommandlockouts                              

\usepackage{cite}
\usepackage{amsmath,amssymb,amsfonts}
\usepackage{algorithmic}
\usepackage{graphicx}
\usepackage{textcomp}
\usepackage{xcolor}
\usepackage{float}

\usepackage{microtype}
\usepackage{subcaption}
\usepackage{booktabs} 
\usepackage{multirow}
\usepackage{pifont}
\newcommand{\cmark}{{\ding{51}}}
\newcommand{\xmark}{{\ding{55}}}
\usepackage{hyperref}
\usepackage{colortbl}
\usepackage{hyperref}

\definecolor{LightCyan}{rgb}{0.88,1,1}
\definecolor{Gray}{gray}{0.9}

\usepackage[latin1]{inputenc}
\usepackage{tikz}
\usetikzlibrary{shapes,arrows}

\tikzstyle{decision} = [diamond, draw, fill=blue!20, 
    text width=4.5em, text badly centered, node distance=3cm, inner sep=0pt]
\tikzstyle{block} = [rectangle, draw, fill=blue!20, 
    text width=5em, text centered, rounded corners, minimum height=4em]
\tikzstyle{line} = [draw, -latex']
\tikzstyle{cloud} = [draw, ellipse,fill=red!20, node distance=3cm,
    minimum height=2em]

\title{\LARGE \bf
OpenLORIS-Object: A Robotic Vision Dataset and Benchmark for \\ Lifelong Deep Learning
}

\author{Qi She$^{1}$, Fan Feng$^{2}$, Xinyue Hao$^{3,4}$, Qihan Yang$^{2}$, Chuanlin Lan$^{2,5}$, Vincenzo Lomonaco$^{6}$, Xuesong Shi$^{1}$, \\ Zhengwei Wang$^{7}$, Yao Guo$^{8}$, Yimin Zhang$^{1}$, Fei Qiao$^{3}$, and Rosa H. M. Chan$^{2}$
\thanks{$^{1}$ Robot Innovation Lab, Intel Labs, Beijing, China}
\thanks{$^{2}$ Department of Electrical Engineering, City University of Hong Kong, China}%
\thanks{$^{3}$ Department of Electronic Engineering, Tsinghua University, China}
\thanks{$^{4}$ Beijing University of Posts and Telecommunications, China}
\thanks{$^{5}$ School of Electronic Information, Wuhan University, China}
\thanks{$^{6}$ Department of Computer Science and Engineering, University of Bologna, Bologna, Italy}
\thanks{$^{7}$ Insight Centre for Data Analytics, Dublin City University, Ireland}
\thanks{$^{8}$ The Hamlyn Centre for Robotic Surgery, Imperial College London, UK }
\thanks{Corresponding author: qi.she@intel.com, rosachan@cityu.edu.hk}
}
\begin{document}

\maketitle
\thispagestyle{empty}
\pagestyle{empty}

\begin{abstract}
The recent breakthroughs in computer vision have benefited from the availability of large representative datasets (e.g. ImageNet and COCO) for training. Yet, robotic vision poses unique challenges for applying visual algorithms developed from these standard computer vision datasets due to their implicit assumption over non-varying distributions for a fixed set of tasks. Fully retraining models each time a new task becomes available is infeasible due to computational, storage and sometimes privacy issues, while na\"{i}ve incremental strategies have been shown to suffer from catastrophic forgetting. It is crucial for the robots to operate continuously under open-set and detrimental conditions with adaptive visual perceptual systems, where lifelong learning is a fundamental capability. However, very few datasets and benchmarks are available to evaluate and compare emerging techniques. To fill this gap, we provide a new lifelong robotic vision dataset (``OpenLORIS-Object") collected via RGB-D cameras. The dataset embeds the challenges faced by a robot in the real-life application and provides new benchmarks for validating lifelong object recognition algorithms. Moreover, we have provided a testbed of $9$ state-of-the-art lifelong learning algorithms. Each of them involves $48$ tasks with $4$ evaluation metrics over the OpenLORIS-Object dataset. The results demonstrate that the object recognition task in the ever-changing difficulty environments is far from being solved and the bottlenecks are at the forward/backward transfer designs. Our dataset and benchmark are publicly available at at \href{https://lifelong-robotic-vision.github.io/dataset/object}{\underline{https://lifelong-robotic-vision.github.io/dataset/object}}.
\end{abstract}

\section{INTRODUCTION}
Humans have the remarkable ability to learn continuously from external environments and inner experiences. One of the grand goals of robots is building an artificial ``lifelong learning'' agent that can shape a cultivated understanding of the world from the current scene and their previous knowledge via an autonomous lifelong development. Recent advances in computer vision and deep learning techniques have been achieved through the occurrence of large-scale datasets, such as ImageNet~\cite{deng2009imagenet} and COCO~\cite{lin2014microsoft}. The breakthroughs in object classification, detection, and segmentation heavily depend on the availability of these large representative datasets for training. However, robotic vision poses new challenges for applying visual algorithms developed from computer vision datasets in real-world applications, due to their implicit assumptions over non-varying distributions for a fixed set of categories and tasks. In practice, the deployed model cannot support the capability to learn and adapt as new data comes in autonomously. 
\begin{figure}[H]
    \centering
    \includegraphics[width=0.9\linewidth]{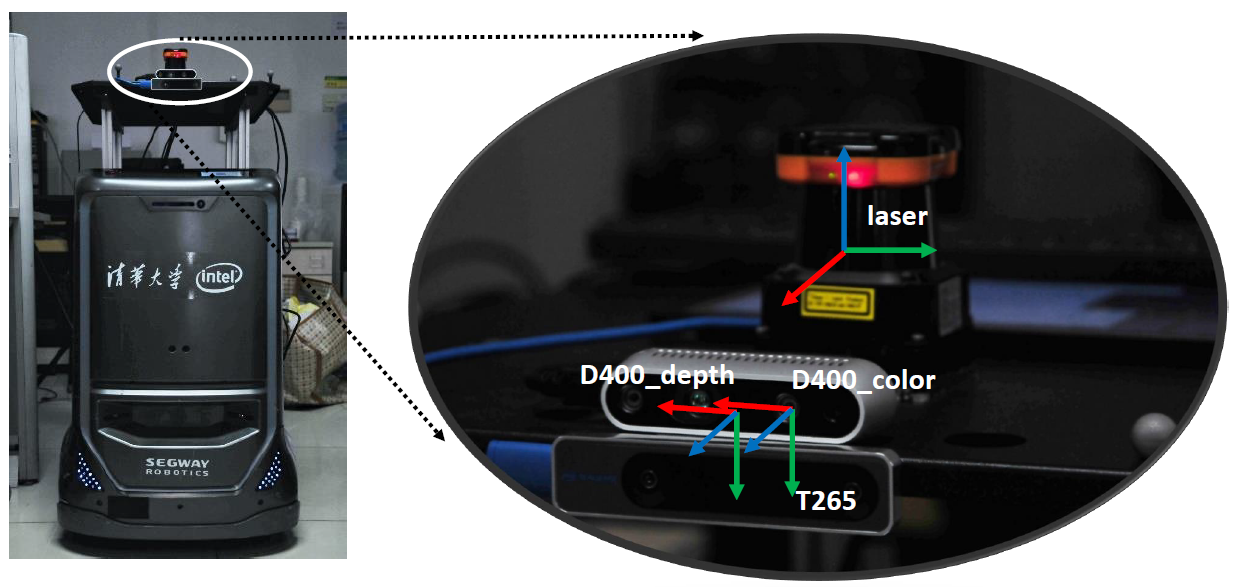}
    \caption{OpenLORIS robotic platform (left) mounted with multiple sensors (right). In OpenLORIS-Object dataset, the RGB-D data is collected from the depth camera.}\vspace{-2mm}
    \label{fig:robot}
\end{figure}

The semantic concepts of the real environment are dynamically changing over time. In real scenarios, robots should be able to operate continuously under open-set and sometimes detrimental conditions, which requires the lifelong learning capability with reliable uncertainty estimates and robust algorithm designs~\cite{sunderhauf2018limits}. \textit{Providing a robotic vision dataset collected from the time-varying environments can accelerate both research and applications of visual models for robotics}. The ideal dataset should contain the objects recorded from the high-variant environments, e.g., the variations of the illumination and clutter. The presence of temporal coherent sessions (i.e., videos where the robot mounted with the camera gently move around the objects) is another key feature since temporal smoothness can be used to simplify object detection, improve classification accuracy and to address unsupervised scenarios~\cite{maltoni2016semi}. In this work, we utilize a real robot mounted with multiple high-resolution sensors (e.g., depth and IMU, see Fig.~\ref{fig:robot}) to actively collect the data from the real-world objects in several kinds of typical scenarios, like homes, offices, and malls. We consider the variations of illumination, occlusion of the objects, object size, camera-object distance/angle, and clutter. These real-world factors encountered by the robots after deployment are explored and quantified at different difficulty levels. Specifically, \textit{we quantify the learning capability of the robotic vision system when faced with the objects appearing in the dynamic environments.} 

When evaluating the existing algorithms designed for real-world lifelong learning, utilizing the whole historical data for retraining the model (i.e., cumulative approach) is unlikely for application engineering because this kind of method needs to 1) store all data streams; and 2) retrain the model with the whole dataset once the new datum is available. In the real-world deployment, it is required to update the trained model with the new data in terms of computation~\cite{al2015efficient}, memory constraint~\cite{lopez2017gradient} or privacy issues (cannot have access to or store previous data)~\cite{mohassel2017secureml}. The vanilla learning approach is transfer learning or fine-tuning method to apply previously learned knowledge on quite similar domain tasks~\cite{pan2009survey}. Yet, they cannot solve the concept drift problem when encountering dissimilar domain dataset~\cite{khamassi2018discussion} and have degradation performances on the previous tasks after learning the new task due to the catastrophic forgetting problem~\cite{goodfellow2013empirical,kirkpatrick2017overcoming,schwarz2018progress}. 

For quantifying the lifelong learning capability of robotic vision systems, three-fold contributions are made in this paper:
\begin{itemize}
\item We have provided a novel RGB-D object dataset for ($\mathbf{L}$)ifel($\mathbf{O}$)ng ($\mathbf{R}$)obotic V($\mathbf{IS}$)ion research called OpenLORIS-object. The dataset is collected via depth cameras. It collects the data under dynamic environments with diverse illumination, occlusion, object size, camera-object distance/angle, and clutter.
    \item We have released benchmarks for evaluating lifelong learning capability of the robotic vision with {\it{ever-changing difficulty}}.
    \item We have done comprehensive analysis with $9$ state-of-the-art lifelong learning algorithms, each of which is evaluated with $48$ recognition tasks on OpenLORIS-Object. It demonstrated the bottlenecks of the SOTAs for learning continuously in a real-life scenario.
\end{itemize}

\section{Related Work}
\subsection{Lifelong Learning Algorithms}
The ultimate lifelong robotic vision system should embed five capabilities: 1) learn the new knowledge and further summarize the patterns from the data; 2) avoid catastrophic forgetting and keep the memory of the old knowledge, especially for the widely-used deep learning techniques; 3) generalize and adapt well to the future time-varying distributions. Traditionally supervised classifiers are trained on one distribution and often fails when faced with a quite different distribution, and most of the current adaptation algorithms may suffer from the performance degradation; 4) be equipped with the few-shot/zero-shot learning capability that can learn from the small dataset; and 5) be able to learn from a theoretically-infinite stream of examples using limited time and memory. In this work, we focus on evaluating the existing algorithms for enabling the \textit{first three capabilities} on the lifelong object recognition task.  
We note that the first three points can be viewed as evaluating the performances of the algorithms over current, previous, and future tasks, respectively. However, a well-known constraint called stability-plasticity dilemma~\cite{mermillod2013stability} impedes this continual adaptation of learning systems. Learning in a parallel and distributed system needs plasticity for integrating the new knowledge, but also stability to prevent forgetting of the previous knowledge. Too much plasticity leads to the learned patterns of previous data being forgotten, whereas too much stability cannot make efficient encoding of the new knowledge. Thus, the recent deep neural network-based algorithms are trying to achieve a trade-off between stability and plasticity when learning from the high-dimensional data space (e.g., image) continuously. 

Conceptually, these approaches can be divided into 1) methods that retrain the whole network via regularizing the model parameters learned from previous tasks, e.g., Learning without Forgetting (LwF)~\cite{li2017learning}, Elastic Weight Consolidation (EWC)~\cite{kirkpatrick2017overcoming} and Synaptic Intelligence (SI)~\cite{zenke2017continual}; 2) methods that dynamically expand/adjust the network architecture if learning new tasks, e.g., Context-dependent Gating (XdG)~\cite{masse2018alleviating} and Dynamic Expandable Network (DEN)~\cite{yoon2017lifelong}; 3) rehearsal approaches gather all methods that save raw samples as memory of past tasks. These samples are used to maintain knowledge about the past in the model and then replayed with samples drawn from the new task when training the model, e.g., Incremental Classifier and Representation Learning (ICaRL)~\cite{rebuffi2017icarl}; and generative replay approaches train generative models on the data distribution, and they are able to afterward sample data from experience when learning new data, e.g., Deep Generative Replay (DGR)~\cite{shin2017continual}, DGR with dual memory~\cite{kamra2017deep} and feedback~\cite{van2018generative}. Most of current generative models are based on standard Generative adversarial networks and its extensions~\cite{goodfellow2014generative,wang2019generative}. For robotic vision, ideally, the lifelong learning should be triggered by the availability of short videos of single objects and performed online on the hardware with fine-grained updates, while the mainstream of methods we study are limited with much lower temporal precision as our previous sequential learning models~\cite{she2018reduced,she2019neural}. 

For evaluating these algorithms, standard datasets like MNIST~\cite{lecun1998gradient} and CUB-200~\cite{welinder2010caltech} are normally utilized. However, the visual algorithms developed from these computer vision datasets have not concerned about learning with every-changing difficulty. They simplify the lifelong learning algorithms with recognizing new object instances or classes under very constraint environment. Moreover, current robotic vision datasets also have limitations on evaluating lifelong learning algorithms because they either neglect some crucial challenges in the environment or have not explicitly quantified the difficulty levels of these challenges in the dataset. Thus, for pushing the boundary of practical lifelong object recognition algorithms, it is required the dataset considers the real-world challenges that the robot encounters and formulate the benchmark as a testbed of object recognition algorithms.
\begin{table*}[t!]
\centering
\resizebox{\textwidth}{!}{

\begin{tabular}{c|c|c|c|c|c|c|c}
\toprule
Dataset & Illumination & Occlusion &  Dimension (pixel)  & Clutter &  Context & Quantifiable & Acquisition \\ \hline

COIL-100~\cite{nene1996columbia} & normal & no & 30-200 & simple & home &  \multirow{9}{*}{\xmark} & turntable \\ 

NORB~\cite{lecun2004learning}  & weak/normal/strong  & no   & \textless 30, 30-200 & simple & outdoor & & turntable \\ 

Oxford Flowers \cite{nilsback2008automated}  & weak/normal & no & 200  & simple & outdoor &  & websites \\


CIFAR-100~\cite{krizhevsky2009learning}  & normal & no & \textgreater 200 & simple & outdoor& &  websites \\ 

UCB-200 \cite{WahCUB_200_2011} &      normal & few & 200 & simple & outdoor & & websites \\ 
                                                           
ROD~\cite{lai2011large}  & normal  & no  & few \textless 30, 30-200  & regular & home & & turntable \\ 

CORe50~\cite{lomonaco2017CORe50}  & normal & no & 30-200 & simple & home/outdoor & & hand hold\\ 

ARID~\cite{arid} & ~80\% weak/normal  & few    & 30-200   & regular/complex & home & & robot\\  \hline

\textbf{Ours: OpenLORIS-Object}    & weak/normal/strong  & no/25\%/50\%  & \textless 30, 30-200, \textgreater 200 & simple/regular/complex & home/office/mall & \cmark & robot \\  \bottomrule     
\end{tabular}} 
\caption{OpenLORIS-Object compared with other object recognition datasets. This summary of the characteristics of different datasets focuses on the variations in illumination, occlusion, object dimension (pixel size) in the image, clutter, context information, and whether or not these characteristics are provided in an explicit (that can be \textbf{quantified}) or an implicit way (cannot isolate
these characteristics of the data, and define the difficulty levels explicitly. Thus we cannot identify how lifelong object recognition algorithms perform w.r.t. the real world challenges rigorously). 
}\vspace{-3mm}
\label{Comparison}
\end{table*}

\subsection{Related Datasets}
One of the relevant robotic vision datasets is RGB-D Object Dataset (ROD)~\cite{lai2011large}, which has become the standard benchmarks in the robotics community for the object recognition task. Although the dataset is well organized and contains over $300$ everyday household objects, it has been acquired under a very constrained setting and neglects some crucial challenges that a robot faces in the real-world deployment. 

Another recently proposed dataset is Autonomous Robot Indoor Dataset (ARID)~\cite{arid}, and the data is collected from a robot patrolling in a defined human environment. Analogously to ROD, the object instances in ARID are organized into various categories. The dataset is manually designed to include real-world characteristics such as variation in lighting conditions, object scale, and background as well as occlusion and clutter. ARID seems to be similar to our dataset, both of which are considering the real-world challenges (e.g., illumination, occlusion) that the robots many naturally encounter; however, two main differences exist. First, OpenLORIS-object has rigorously isolated each characteristic/environment factor of the dataset, such as illumination, occlusion, object pixel size, and clutter, and defines the difficulty levels of each factor explicitly. However, ARID contains these challenges in an implicit manner. Although these challenges exist implicitly in a real-life deployment, during system design and development stages, the dataset including implicit variants cannot benefit the system as much as ours via providing the difficulty levels of each challenge explicitly; Second, the OpenLORIS-Object is designed for evaluating the lifelong learning capability of the robotic vision system. Thus we have provided the benchmarks for several ``ever-changing difficulty" scenarios.

Continual Object Recognition Dataset (CORe50)~\cite{lomonaco2017CORe50} is a collection of $50$ domestic objects, which evaluates the continual learning capability of the models. Different from OpenLORIS-Object, they focus on incrementally recognizing new instances or new classes. We focus on how to learn the objects under varying environmental conditions, which is essential for enabling the robots to perform continuously and robustly in the dynamic environment. Moreover, Objects in CORe50 are handhold by the operator, and the camera point-of-view is that of the operator's eyes. While OpenLORIS-Object is more suitable for autonomous system development because the data is acquired via the real robots mounted with depth cameras, which is in an active vision manner.

Non-I.I.D. Image dataset with Contexts (NICO) supports testing the lifelong learning algorithms on the Non-I.I.D. data, which focuses on the context variants of the same object. We highlight that NICO is the mixed effects of our considered factors. Furthermore, we have decomposed $4$ orthogonal contexts, e.g., illumination, occlusion, object pixel sizes, clutter in OpenLORIS-Object. Note that the context in our dataset is classified as home, office and shopping mall scenarios, and we admit that they also contain mixed factors as NICO (we tried to keep other factors at the normal level). More robotic vision datasets exists in lifelong research~\cite{shi2019we}, but in this work, the related datasets compared focus on object recognition problem. 

We briefly show the features of OpenLORIS-Object compared with others in Table~\ref{Comparison}. It demonstrates that ours is quantifiable and more complete w.r.t. the real-life challenges for robotic object recognition. 

\begin{table*}[t!]
    \centering
    \begin{tabular}{c|c|c|c|c|c|c|c|c}
    \toprule
         Level & Illumination & Occlusion (percentage) & Object Pixel Size (pixels) & Clutter & Context & \#Classes & \#Instances  & \#Scenes\\ \hline
         1   &  Strong & $0\%$ & $> 200 \times 200$ & Simple & \multirow{3}{*}{Home/office/mall} & \multirow{3}{*}{40} & \multirow{3}{*}{121} & \multirow{3}{*}{20} \\ 
         2  &  Normal      & $25\%$ & $30 \times 30-200\times 200$ & Normal & & &\\ 
         3 & Weak    & $50\%$      &   $< 30 \times 30$ & Complex & & &\\
    \bottomrule
    \end{tabular}
    \caption{Details of each $3$ levels for $4$ real-life robotic vision challenges.}
    \label{Task_level}
\end{table*}
\begin{table*}[t!]
\centering
\begin{tabular}{c|c|c|c}
\toprule
Accuracy & BWT & FWT & Over-all accuracy  \\ \hline 
{\footnotesize$\sum_{{i}\ge{j}}^{N}R_{i,j}/\dfrac{N(N+1)}{2}$} & 
{\footnotesize$\sum_{i>j}^{N}R_{i,j}/\dfrac{N(N-1)}{2}$} &
  {\footnotesize$\sum_{i<j}^{N}R_{i,j}/\dfrac{N(N-1)}{2}$} &
{\footnotesize$\sum^{N}_{i,j}R_{i,j}/{N^{2}}$}\\ \bottomrule
\end{tabular}
\caption{Four evaluation metrics for the lifelong object recognition task.}
\label{tab:metrics}
\end{table*}
\section{OpenLORIS-Object Dataset}
\subsection{Dataset Collection}
Several grounded robots mounted by depth cameras and other sensors are used for the data collection. These robots are moving in the offices, homes, and malls, where the scenes are diverse and changing all the time. In the OpenLORIS-Object dataset, we provide the RGB-D video dataset for the objects.

\subsection{Dataset Details}
We include the common challenges that the robot is usually faced with, such as illumination, occlusion, camera-object distance, etc. Furthermore, we explicitly decompose these factors from real-life environments and have quantified their difficulty levels. In summary, to better understand which characteristics of robotic data negatively influence the results of the lifelong object recognition, we independently consider: 1) illumination, 2) occlusion, 3) object size, 4) camera-object distance, 5) camera-object angle, and 6) clutter. 
\begin{itemize}
    \item[1).]\textbf{Illumination}. The illumination can vary significantly across time, e.g., day and night. We repeat the data collection under weak, normal, and strong lighting conditions, respectively. The task becomes challenging with lights to be very weak.
    \item[2).]\textbf{Occlusion}. Occlusion happens when a part of an object is hidden by other objects, or only a portion of the object is visible in the field of view. Occlusion significantly increases the difficulty for recognition.
    \item[3).]\textbf{Object size}. Small-size objects make the task challenging, like dry batteries or glue sticks.
    \item[4).]\textbf{Camera-object distance}. It affects actual pixels of the objects in the image.
    \item[5).]\textbf{Camera-object angle}. The angles between the cameras and objects affect the attributes detected from the object.
    \item[6).]\textbf{Clutter}. The presence of other objects in the vicinity of the considered object may interfere with the classification task.
\end{itemize}

The $1^{st}$ version of OpenLORIS-Object is a collection of $121$ instances, including $40$ categories daily necessities objects under $20$ scenes. For each instance, a $17$ to $25$ seconds video (at $30$ fps) has been recorded with a depth camera delivering around $500$ to $750$ frames ($260$ to $600$ distinguishable object views are manually picked and provided in the dataset). $4$ environmental factors, each has $3$ level changes, are considered explicitly, including illumination variants during recording, occlusion percentage of the objects, object pixel size in each frame, and the clutter of the scene. Note that the variables of 3) object size and 4) camera-object distance are combined together because in the real-world scenarios, it is hard to distinguish the effects of these two factors brought to the actual data collected from the mobile robots, but we can identify their joint effects on the actual pixel sizes of the objects in the frames roughly. The variable 5) is considered as different recorded views of the objects. The defined three difficulty levels for each factor are shown in Table.~\ref{Task_level} (totally we have $12$ levels w.r.t. the environment factors across all instances). The levels $1$, $2$, and $3$ are ranked with increasing difficulties. 

\begin{figure}[H]
\centering
  \includegraphics[width=\linewidth]{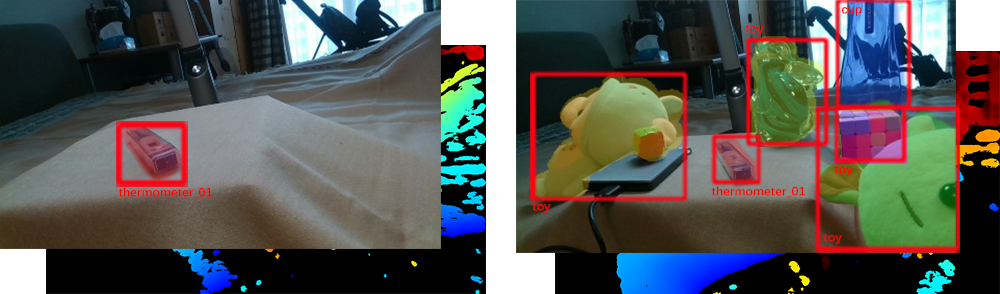}
  \caption{Example of two RGB-D frames of simple clutter (left) and complex clutter (right) from OpenLORIS-Object Dataset with 2D bounding box and mask annotations. 
  }
  \label{fig:object_bb}
\end{figure}

\begin{figure*}[t!]
\centering
  \includegraphics[width=1\linewidth]{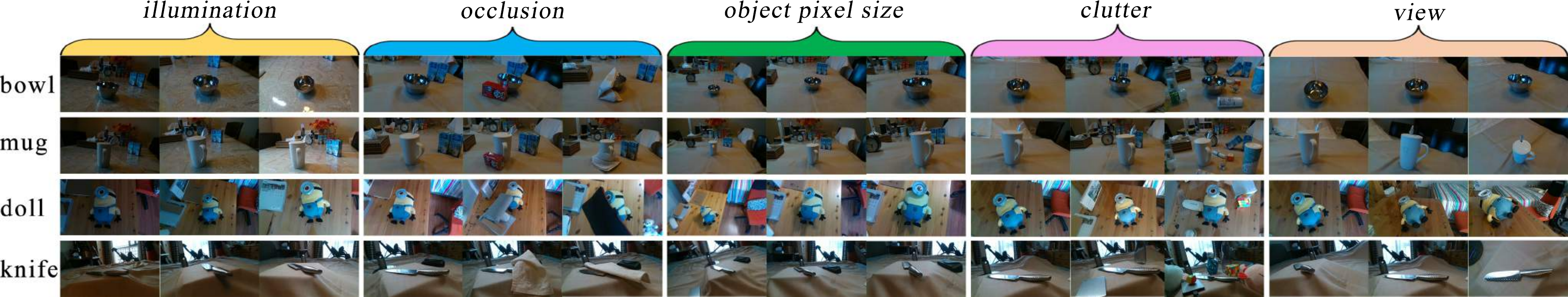}
  \caption{Picked samples of $4$ objects (row) under multiple level environment conditions (column). The variants from left to right are illumination (weak, normal, and strong); occlusion ($0\%$, $25\%$, and $50\%$); object pixel size ($<30 \times 30$, $30 \times 30 - 200 \times 200$, and $ > 200 \times 200$); clutter (simple, normal and complex); and multi-views of the objects. (Note that we use different views as training samples of each difficulty level in each factor). \vspace{-2mm}
  }
  \label{fig:object}
\end{figure*}

For each instance at each level, we provided $260$ to $600$ samples, both have RGB and depth images. Thus, the total images provided is around $2$ (RGB and depth) $\times$ $381$ (mean samples per instance)$\times$ $121$ (instances) $\times$ $4$ (factors per level) $\times$ $3$ (difficulty levels) = $1,106,424$ images. Also, we have provided bounding boxes and masks for each RGB image. An example of two RGB-D frames of simple and complex clutter with 2D bounding box and mask annotations is shown in Fig.~\ref{fig:object_bb}. The size of images under illumination, occlusion and clutter factors is $424 \times 240$ pixels, and the size of images under object pixel size factor are $424 \times 240$, $320 \times 180$, $1280 \times 720$ pixels. Picked samples have been shown in Fig.~\ref{fig:object}. 

\section{Experiments: Lifelong Object Recognition with Ever-changing Difficulty}
\subsection{State-of-the-art Methods}
We evaluate $9$ lifelong learning algorithms over OpenLORIS-Object dataset. These methods can be classified into $3$ categories: 1) transfer and multi-task learning: na\"{i}ve and cumulative methods~\cite{lomonaco2017CORe50, gao2019nddr}; 2) regularization approaches: Learning without Forgetting (LwF)~\cite{li2017learning}, Elastic Weight Consolidation (EWC)~\cite{kirkpatrick2017overcoming}, Online EWC~\cite{schwarz2018progress} and Synaptic Intelligence (SI)~\cite{zenke2017continual}; and 3) Generative Replay approaches: Deep Generative Replay (DGR)~\cite{shin2017continual}, DGR with distillation~\cite{venkatesan2017strategy,wu2018incremental} and DGR with feedback~\cite{van2018generative}. More details of these methods can be seen in the recent review~\cite{parisi2019continual}. 

\subsection{Evaluation Metrics}
Finding the metrics, which are most useful to report the Lifelong learning performances, are non-trivial like investigating how to evaluate other deep learning methods~\cite{wang2019neuroscore}. In this paper, for a comprehensive analysis of SOTAs over OpenLORIS-Object datasets, quite different from ROD~\cite{lai2011large}, CORe50~\cite{lomonaco2017CORe50}, ARID~\cite{arid}, and NICO~\cite{he2019nico}, we utilize four metrics for evaluating the performances~\cite{diaz2018don}: Accuracy, Backward transfer (BWT), Forward transfer (FWT), and Over-all accuracy as shown in Table~\ref{tab:metrics}. It is denoted that these metrics still focus on the accuracy aspect for learning, while ignoring the key indicators of computational efficiency and memory storage.

During lifelong learning process, the data $\mathcal{D}$ is a potentially infinite $\textit{sequence}$ of unknown distributions $\mathcal{D} = \{D_{1},\cdots,D_{N}\}$. For the dataset $D_{n}$, We denote the training set $Tr_{n}$ and testing set $Te_{n}$, and define the task $T_{n}$ for recognizing the object categories in this dataset. A train-test accuracy matrix $R \in \mathcal{R}^{N\times N}$ contains in each entry $R_{ij}$ the testing classification accuracy of dataset $Te_{j}$ after training the model over the dataset $Tr_{i}$, which is shown in Table.~\ref{tab:train-test-matrix}. 
\begin{table}[H]
\centering
\begin{tabular}{c|cccc}
\toprule
 $R$   & $Te_{1}$ & $Te_{2}$ & $\cdots$ & $Te_{N}$ \\ \hline
$Tr_{1}$ & $R_{11}$ & \cellcolor{LightCyan}$R_{12}$ & \cellcolor{LightCyan} $\cdots$ &  \cellcolor{LightCyan}$R_{1N}$ \\
$Tr_{2}$ & \cellcolor{Gray}$R_{21}$ & $R_{22}$ &  \cellcolor{LightCyan}$\cdots$ &  \cellcolor{LightCyan}$R_{2N}$ \\
$\cdots$ & \cellcolor{Gray}$\cdots$ & \cellcolor{Gray}$\cdots$ & $\cdots$ &  \cellcolor{LightCyan}$\cdots$ \\ 
$Tr_{N}$ & \cellcolor{Gray} $R_{N1}$ & \cellcolor{Gray}$R_{N2}$ & \cellcolor{Gray}$\cdots$  & $R_{NN}$ \\ \bottomrule
\end{tabular}
\caption{Train-test accuracy matrix $R$, where $Tr =$ training data, $Te =$ testing data, and $R_{ij} = $ classification accuracy of the model training on $Tr_{i}$ and testing on $Te_{j}$. The number of tasks is $N$, and the train/test split is $8:2$.} \vspace{-2mm}
\label{tab:train-test-matrix}
\end{table}

The Accuracy metric considers the performance of the model at very timestep $i$ in time that can better characterize the dynamics of the learning algorithms (average of white and gray elements in Table~\ref{tab:train-test-matrix}); BWT evaluates the memorization capability of the algorithms, which measures the accuracy over previously encountered tasks (average of gray elements in Table~\ref{tab:train-test-matrix}); FWT measures the influence that learning the current task on the performance of
future tasks (average of cyan elements in Table~\ref{tab:train-test-matrix}); and Over-all accuracy summarizes the performances on all the previous, current, and future tasks, which can be viewed as an overall metric for a specific model.


\subsection{Benchmarks of Lifelong Object Recognition}
\paragraph{Single factor analysis with ever-changing difficulty} 
The experiments are conducted on the $4$ robotic vision challenges, each of which has $3$ difficulty levels that can be explored under the sequential learning settings. We first investigate the individual factor, and change the difficulty levels of each continuously. Note that we keep other factors at the level $1$ as in Table~\ref{Task_level} when investigating each factor, e.g., the levels of the illumination can be weak, normal, and strong, and at the same time, the occlusion is kept at $0\%$, the object pixel size is larger $200\times 200$, and the clutter is simple. 

\begin{figure}[htbp]
\centering
  \includegraphics[width=1\linewidth]{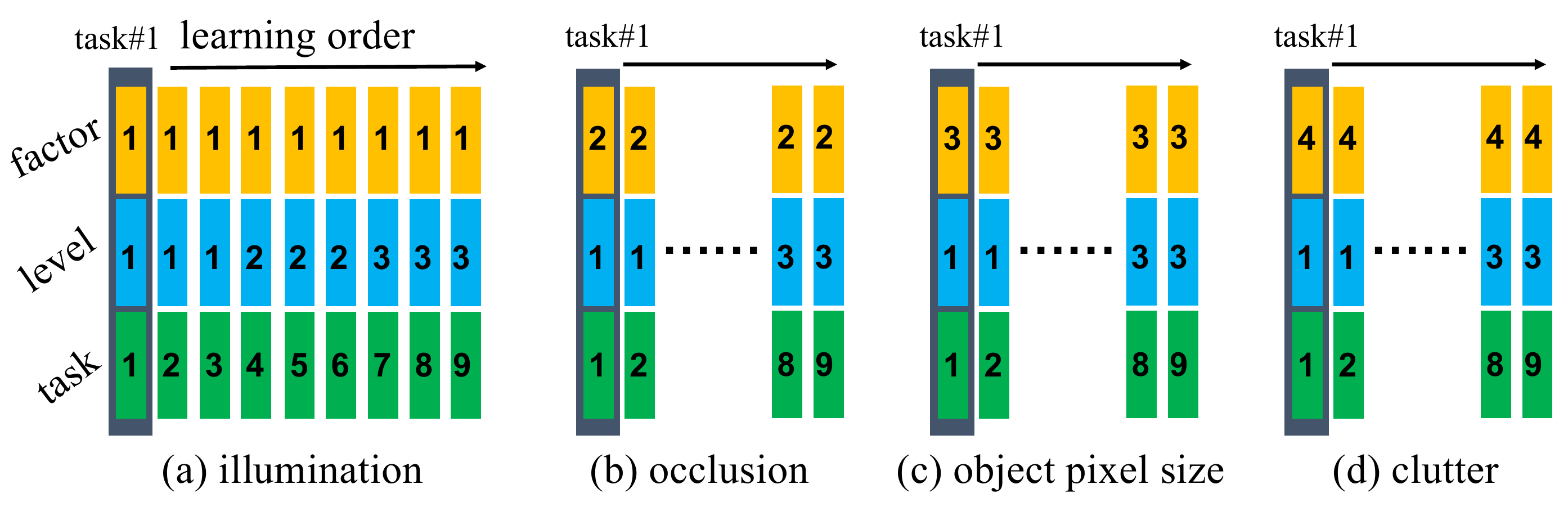}
  \caption{Four-factor analysis (illumination, occlusion, object pixel size, and clutter) under the sequential learning setting. Yellow bars indicate the factor encountered (``1": illumination, ``2'': occlusion, ``3": object pixel size, and ``4": clutter); blue bars highlight the difficulty levels within each factor, and green bars represent the task ID. Within each difficulty level, three tasks are provided w.r.t. their variants in object views.}
  \label{fig:factor}
\end{figure}

Fig.~\ref{fig:factor} demonstrates the experimental details of each factor analysis. For example in the fig.~\ref{fig:factor} (a), under illumination variants (shown in yellow bars ``factor"), the model should be updated with the data from the difficulty level $1$, $2$, and $3$ (shown in blue bars ``level") for totally $9$ tasks (shown in green bars ``task"). We separate each difficulty level into $3$ tasks (e.g., blue bars we have three $1$/$2$/$3$ level) with different views. The same experiment has been done on occlusion, object pixel size, and clutter factors. For each task, the total images are around $15,353$. The number of images in training, testing, validation sets is around $12,283$, $1,535$ and $1,535$ respectively.

\begin{figure*}[t!]
        \begin{subfigure}[b]{0.24\textwidth}
                \includegraphics[width=\linewidth]{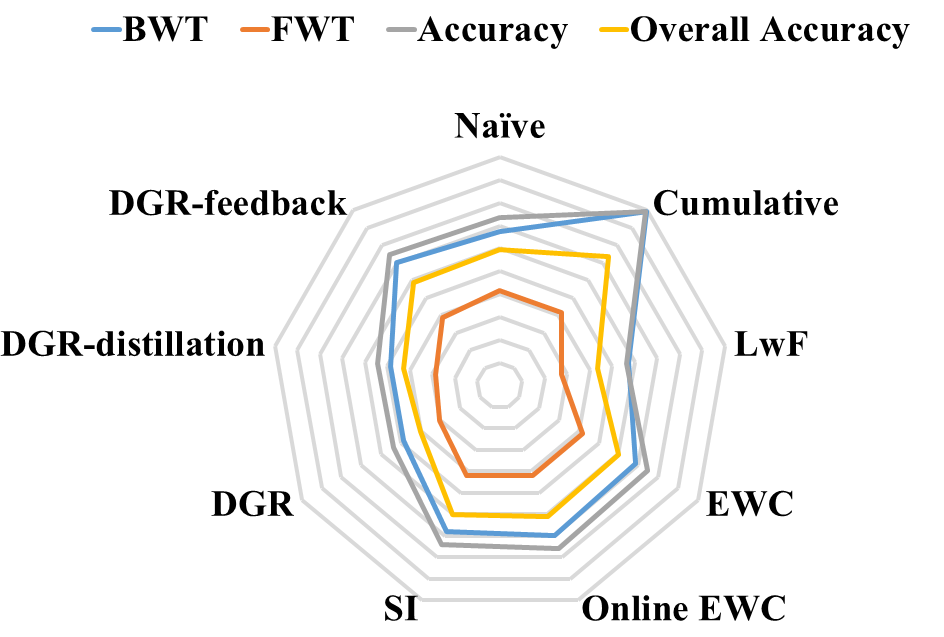}
                \caption{illumination}
                \label{fig:factor1}
        \end{subfigure}
        \begin{subfigure}[b]{0.24\textwidth}
        \includegraphics[width=\linewidth]{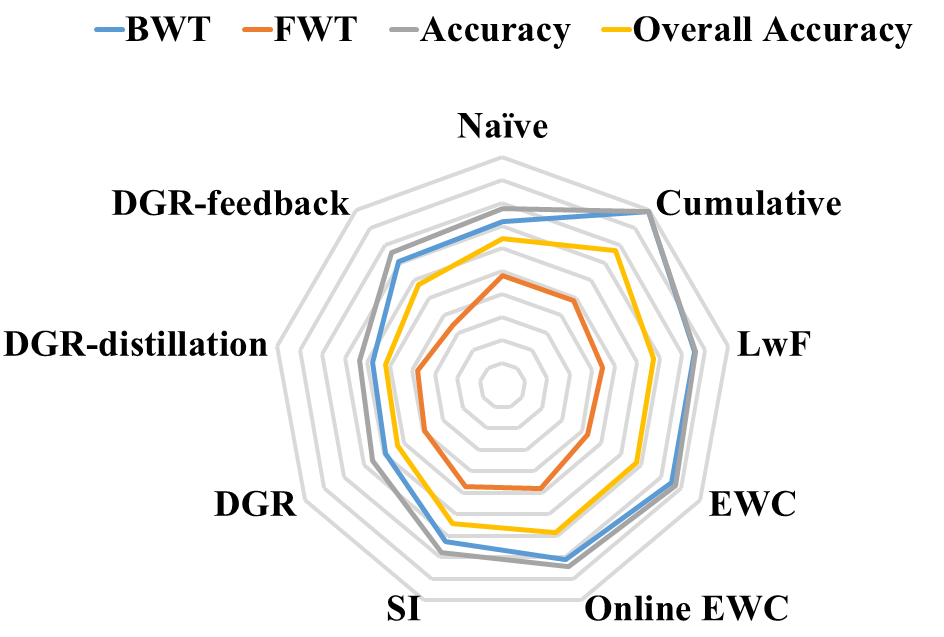}
                \caption{occlusion}
                \label{fig:factor2}
        \end{subfigure}
        \begin{subfigure}[b]{0.24\textwidth}
                \includegraphics[width=\linewidth]{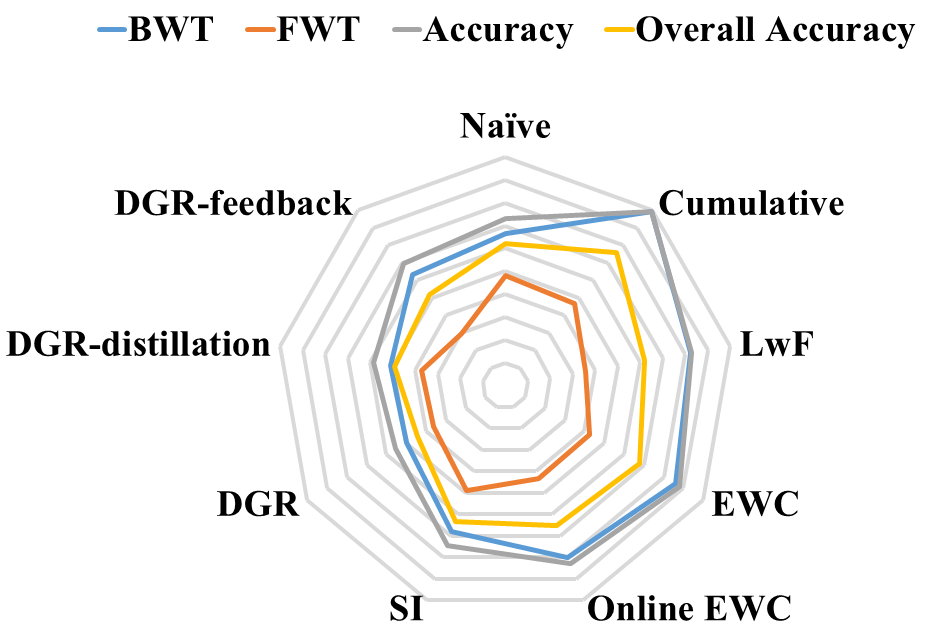}
                \caption{object pixel size}
                \label{fig:factor3}
        \end{subfigure}
        \begin{subfigure}[b]{0.24\textwidth}
                \includegraphics[width=\linewidth]{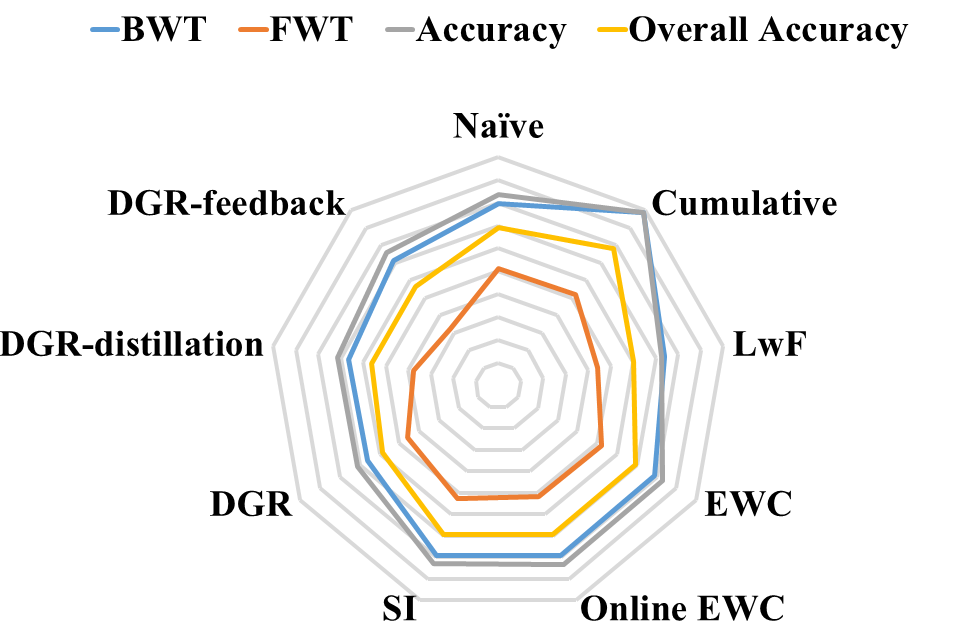}
                \caption{clutter}
                \label{fig:factor4}
        \end{subfigure}
        \caption{The spider chart of $4$ evaluation metrics: Accuracy (grey), BWT (blue), FWT (red), and Over-all accuracy (yellow) of $9$ lifelong learning algorithms, which are evaluated on illumination, occlusion, object pixel size, and clutter factors. Larger area is better. The maximum value of each evaluation metric is $100\%$.}\label{fig:factor_result} \vspace{-5mm}
\end{figure*}

The performances of all $36$ tasks ($4$ factors $\times 9$ tasks/factor) have been evaluated with $4$ metrics (Accuracy, BWT, FWT, and Over-all accuracy) obtained from train-test accuracy matrix as in Table~\ref{tab:train-test-matrix}. In detail, for each factor (including $9$ sequential tasks), we have $1$ train-test accuracy matrix, thus finally we obtain $4$ matrices ($4$ factors) and get the evaluation metrics from them. The results are shown in Fig.~\ref{fig:factor_result}. $9$ commonly-used lifelong learning algorithms are implemented to test the overall performances on OpenLORIS-Object dataset. Compared with the other $3$ factors, the illumination factor is more challenging with smaller areas of all metrics. It also consistently conveys that because of the low accuracy of the forward transfer (FWT) across all methods and factors ($40\%-50\%$ classification accuracy), it will finally lead to lower Over-all accuracy (around $60\%$). 

As is known, most of existing lifelong learning algorithms are designed for overcoming catastrophic forgetting problems. They focus on keeping the backward transfer (BWT) (blue lines) as large as possible, approaching to Accuracy (grey lines), but ignoring the forward transfer capability of the model design. Furthermore, the cumulative approach (retrain with both previous and current data) performs best, however, it needs much more memory storage (linearly grow with the number of tasks encountered), which is impossible for the real-world deployment scenario. More benchmarks for testing the robustness of lifelong learning algorithms with each of randomly encountered challenges can be found in assistive robotics~\cite{feng2019challenges}, which is a more realistic problem that robot may be faced with. 

\begin{figure}[htbp]
\centering
  \includegraphics[width=1\linewidth]{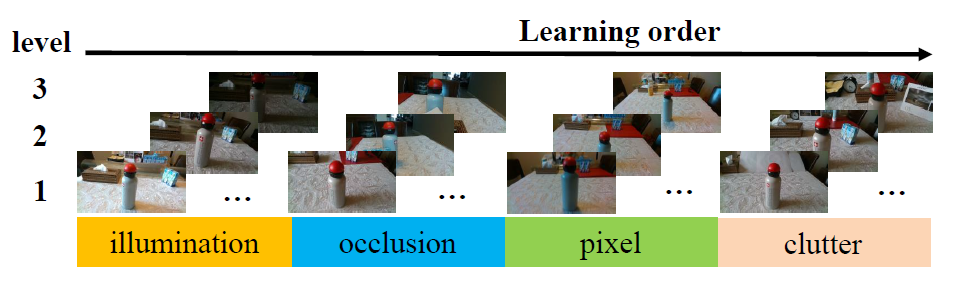}
  \caption{Sequential factors analysis. The models are trained for learning the $4$ factors with $3$ difficulty levels continuously.}
  \label{fig:sefactor} 
  \vspace{-2mm}
\end{figure}

\paragraph{Sequential factors analysis with ever-changing difficulty}
We further explore the task learning capabilities when encountering four factors sequentially.  As shown in Fig.~\ref{fig:sefactor}, the data from $4$ factors with $3$ difficulty levels (totally $12$ tasks) are learned sequentially with about $36,850$ training images and $4,600$ testing images ($121$ objects) for each task. The number of total training and testing images of all the tasks is about $442,194$ and $55,274$. We would like to test the robustness and adaptation capabilities of the lifelong learning algorithms for the long sequential tasks with more variants encountered. 

\begin{figure}[H]
\centering
  \includegraphics[width=0.95\linewidth]{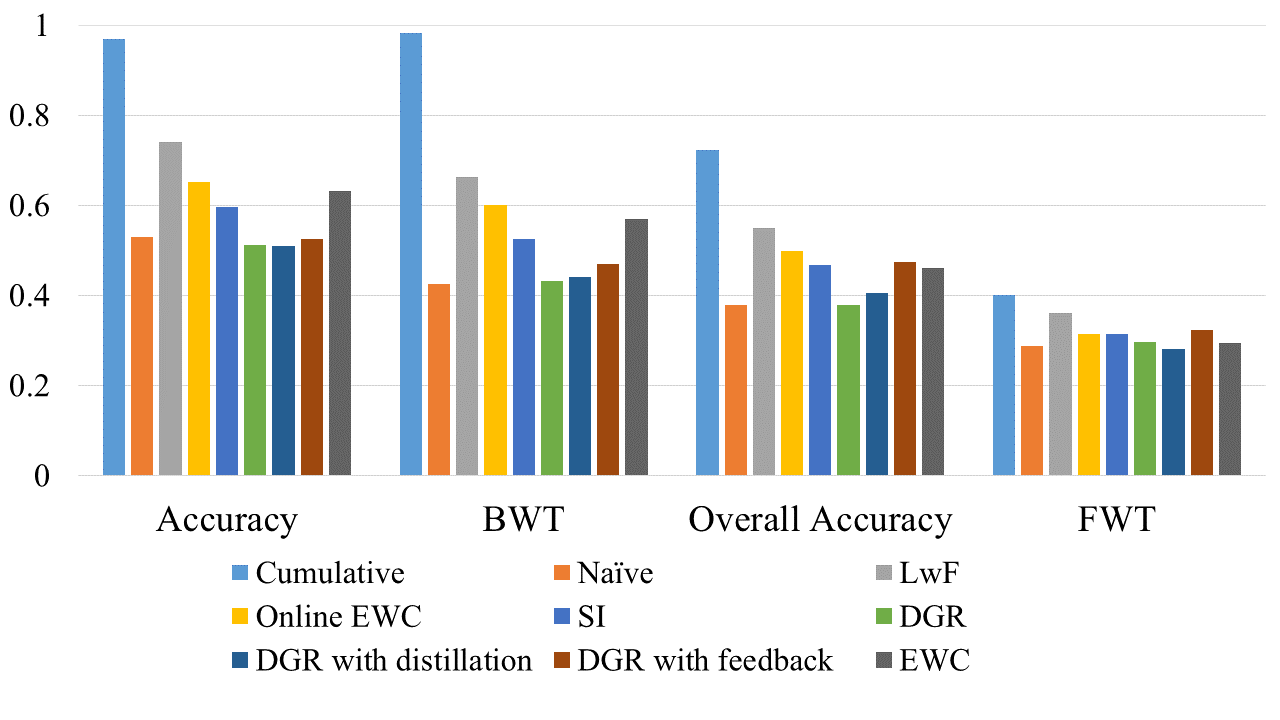}
  \caption{Evaluation results of sequential task learning.}
  \label{fig:sefactor_result}
\end{figure}

The evaluation results of $12$ sequential tasks combining all factors are shown in Fig.~\ref{fig:sefactor_result}. The accuracy, BWT, FWT and overall accuracy degrade significantly compared with single factor analysis. For example, the BWT, FWT and Overall accuracy are averagely below $60\%$ with more high-variant task learning, while in single factor analysis, they can achieve $80\%$ accuracy. It conveys that the existing methods are not stable and robust enough for dealing with long sequential and high variant tasks. 

We would like to denote that the current experiments are conducted with focus on lifelong/continual learning capability, while we have not designed more specific and precise algorithms/modules for overcoming the illumination, occlusion, small-object detection, and object recognition with complex clutter problems. 

\section{Conclusion}
\label{Connclusion}
In order to enable the robotic vision system with lifelong learning capability, we provide a novel lifelong object recognition dataset (``OpenLORIS-Object"). The dataset embeds the real-world challenges (e.g., illumination, occlusion, object pixel sizes, clutter) faced by a robot after deployment and provides novel benchmarks for validating state-of-the-art lifelong learning algorithms. With intensive experiments ($9$ algorithms each of which has $48$ learning tasks), we have found the object recognition task in the ever-changing difficulty (dynamic) environments is far from being solved, which is another challenge besides the object recognition task in the static scene. Two main conclusions can be drawn here:
\begin{itemize}
    \item The bottlenecks of developing robotic perception systems in real-world scenarios are forward and backward transfer model designs, which can figure out how the knowledge can transfer across different scenes.
    \item Under ever-changing difficulty environment, the SOTAs degrade sharply with more tasks, which demonstrates the current learning algorithms are not robust enough, far from being deployed to the real world.
\end{itemize}

\section{Acknowledgement}
\label{acknowlegement}
The work was partially supported by a grant from the Research Grants Council of the Hong Kong Special Administrative Region, China (Project No. CityU 11215618).
The authors would like to thank Hong Pong Ho from Intel RealSense Team for the technical support of RealSense cameras for recording the high-quality RGB-D data sequences. Thank Yang Peng, Kelvin Yu, Dion Gavin Mascarenhas for data collection and labeling. 
\bibliography{root}
\bibliographystyle{IEEEtran}

\end{document}